\documentclass[10pt,twocolumn,letterpaper]{article}

\usepackage[utf8]{inputenc}
\usepackage[T1]{fontenc}

\usepackage{wacv}
\makeatletter
\@namedef{ver@everyshi.sty}{}
\makeatother
\usepackage{tikz}

\usepackage{times}

\usepackage[textsize=scriptsize
,disable
]{todonotes}

\usepackage{tabu}
\usepackage{multirow}
\usepackage{graphicx}
\usepackage{overpic}
\usepackage{array}
\usepackage{wrapfig}
\usepackage{booktabs}
\usepackage{tabu}
\usepackage{subcaption}
\usepackage[font=small]{caption}
\usepackage{blindtext}
\usepackage{amsmath}
\usepackage{nccmath}
\usepackage{url}
\usepackage{bbm}
\usepackage{siunitx,etoolbox}
\robustify\bfseries
\newif\ifarxiv
\arxivtrue %

\ifarxiv
\usepackage{setspace}
\usepackage{fancyhdr}
\newcommand*{\doi}[1]{{\scriptsize doi:\href{http://dx.doi.org/\detokenize{#1}}{\detokenize{#1}}}}

\fi

\usepackage[pagebackref=false,breaklinks=true,letterpaper=true,colorlinks,bookmarks=false]{hyperref}

\usepackage[capitalize]{cleveref}

\wacvfinalcopy %

\def\wacvPaperID{730} %

\ifarxiv\else\ifwacvfinal\fi\fi
\clearpage{}%

\newcommand{\sizethree}[3]{\ensuremath{#1\mkern-2mu\times\mkern-2mu#2\mkern-2mu\times\mkern-2mu#3\xspace}}

\makeatletter
\DeclareRobustCommand\onedot{\futurelet\@let@token\@onedot}
\def\@onedot{\ifx\@let@token.\else.\null\fi\xspace}

\def\eg{\emph{e.g}\onedot} 
\def\ie{\emph{i.e}\onedot} 
\def\cf{\emph{cf.}\xspace}

\def\etal{\emph{et~al}\onedot}
\makeatother

\newcommand{\stardist}{\mbox{\small\textsc{StarDist-3D}}\xspace}
\newcommand{\hmaxima}{\mbox{\small\textsc{IFT-Watershed}}\xspace}
\newcommand{\unet}{\mbox{\small\textsc{U-Net}}\xspace}
\newcommand{\unetws}{\mbox{\small\textsc{U-Net+}}\xspace}

\newcommand{\dataworm}{\mbox{\textsc{Worm}}\xspace}
\newcommand{\datako}{\mbox{\textsc{Parhyale}}\xspace}

\clearpage{}%

\begin{document}

\title{Star-convex Polyhedra for 3D Object Detection and Segmentation in Microscopy}

\author{%
Martin Weigert$^{1,2,3,\star}$
\hfill Uwe Schmidt$^{2,3,\star}$
\hfill Robert Haase$^{2,3}$
\hfill Ko Sugawara$^{4,5}$
\hfill Gene Myers$^{2,3}$
\\[1.0em]
$^{1}$Institute of Bioengineering, École Polytechnique Fédérale de Lausanne (EPFL), Switzerland\\
$^{2}$Max Planck Institute of Molecular Cell Biology and Genetics (MPI-CBG), Dresden, Germany\\
$^{3}$Center for Systems Biology Dresden (CSBD), Germany\\
$^{4}$Institut de Génomique Fonctionnelle de Lyon (IGFL), École Normale Supérieure de Lyon, France\\
$^{5}$Centre National de la Recherche Scientifique (CNRS), Paris, France%
}
\maketitle
\ifwacvfinal\thispagestyle{empty}\fi
\ifarxiv\thispagestyle{fancy}\fi

\let\oldthefootnote\thefootnote
\renewcommand{\thefootnote}{$\star$}
\footnotetext[1]{Equal contribution}
\let\thefootnote\oldthefootnote

\begin{abstract}

Accurate detection and segmentation of cell nuclei in volumetric (3D)
fluorescence microscopy datasets is an important step in many biomedical
research projects.
Although many automated methods for these tasks exist, they often struggle for
images with low signal-to-noise ratios and/or dense packing of nuclei.
It was recently shown for 2D microscopy images
that these issues can be alleviated by training a neural network to directly
predict a suitable shape representation (star-convex polygon) for cell nuclei.
In this paper, we adopt and extend this approach to 3D volumes by using star-convex polyhedra to represent cell nuclei and similar shapes. To that end, we overcome the challenges of
\emph{1)} finding parameter-efficient star-convex polyhedra representations that can faithfully describe cell nuclei shapes,
\emph{2)} adapting to anisotropic voxel sizes often found in fluorescence microscopy datasets, and
\emph{3)} efficiently computing intersections between pairs of star-convex polyhedra (required for non-maximum suppression).
Although our approach is quite general, since star-convex polyhedra include common shapes like bounding boxes and spheres as special cases, our focus is on accurate detection and segmentation of cell nuclei.
Finally, we demonstrate on two challenging datasets that our approach (\stardist) leads to superior results when compared to classical and deep learning based methods.

\end{abstract}

\newcommand{\figOverview}{{
\begin{figure*}[t]
  \centering
  {
    \includegraphics[width=1\linewidth]{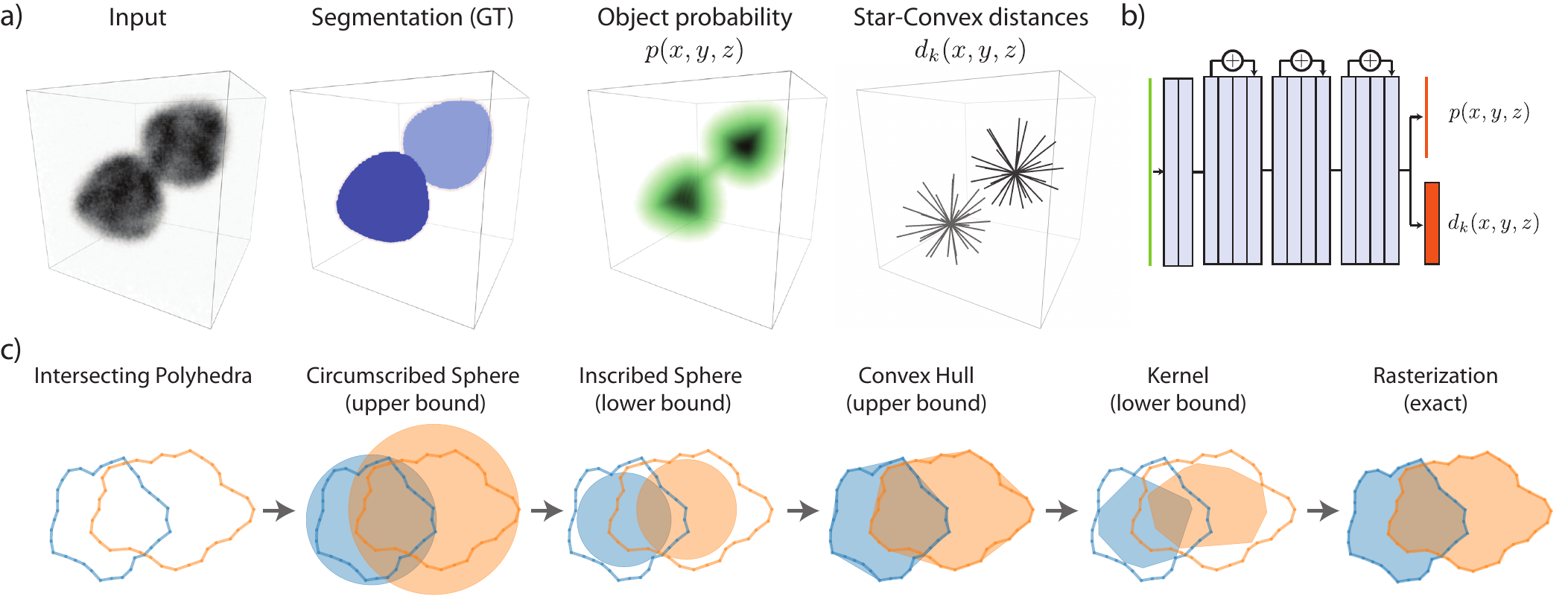}
  }%
  \vspace{-15pt}
\caption{
a)
The proposed \stardist method is trained to densely predict object probabilities $p$
and radial distances $d_k$ to object boundaries.
b) Schematic of our CNN architecture based on ResNet \cite{he2016}.
c) During non-maximum suppression we use successively tighter bounds to efficiently determine if the intersection volume of two star-convex polyhedra is above a given threshold (only shown in 2D here).
}
\label{fig:overview}
\end{figure*}
}}

\newcommand{\figNMS}{{
\begin{figure}[t]
  \centering
  {
    \includegraphics[width=1\textwidth]{figures/intersection.pdf}
  }%
  \vspace{-3pt}

  \caption{blub}

\label{fig:nms}
\end{figure}
}}

\newcommand{\figDatasets}{
    \begin{figure*}[t!]
  {
    \includegraphics[width=1\textwidth]{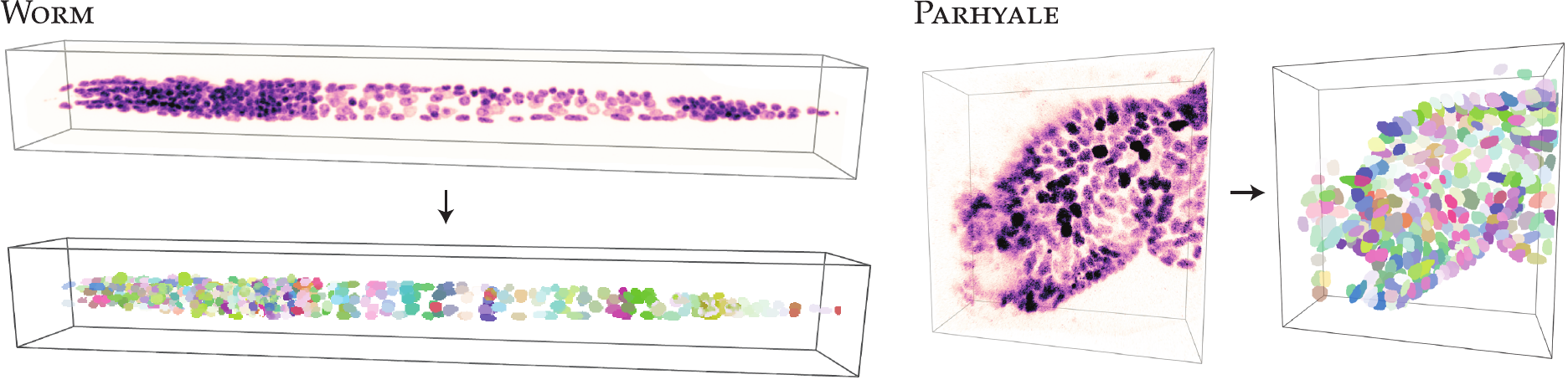}
  }%
  \vspace{-16pt}
\caption{Datasets used in our experiments. Shown are raw input images (purple) and associated ground-truth instance segmentation labels (colored) for a single volume of the \dataworm (left) and \datako (right) datasets.}
\label{fig:datasets}
\end{figure*}
}

\newcommand{\figResults}{{
    \begin{figure*}[t]
      \scriptsize
      a) \dataworm \\ %
      \renewcommand{\arraystretch}{0.5}
      \begin{tabular}{ccc}
    Input & GT & \textsc{IFT-Watershed}\\
      \includegraphics[width=.32\linewidth]{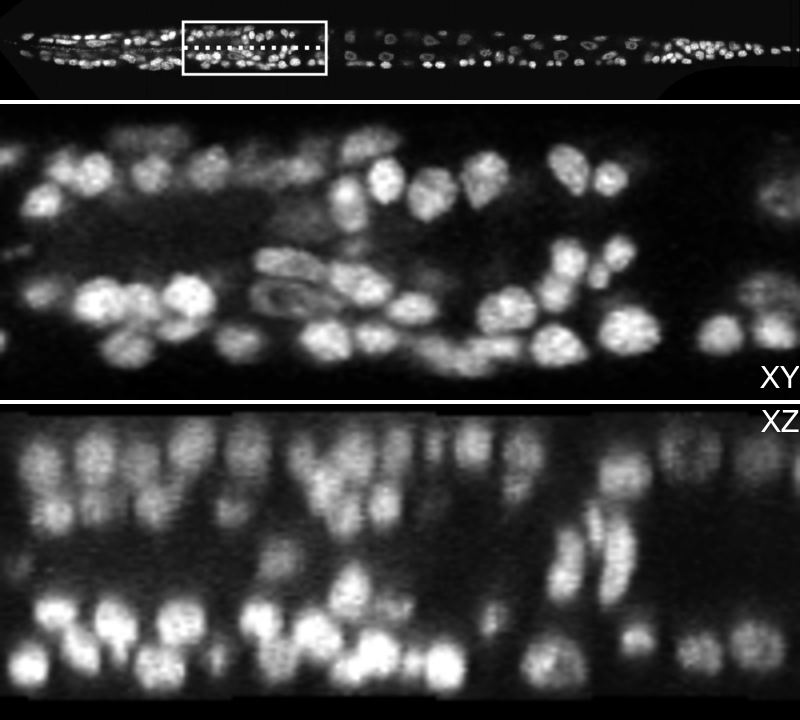}
      &\includegraphics[width=.32\linewidth]{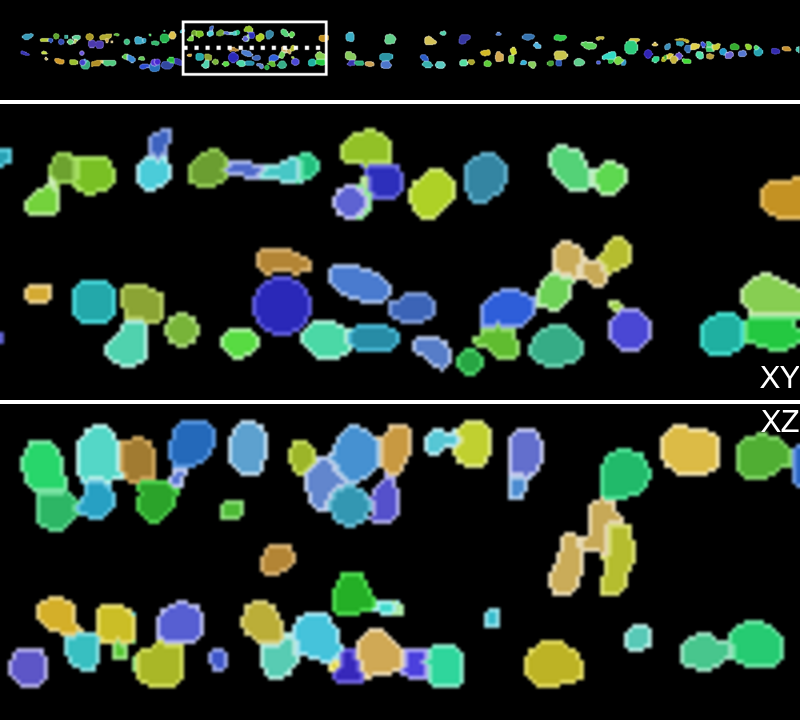}
      &\includegraphics[width=.32\linewidth]{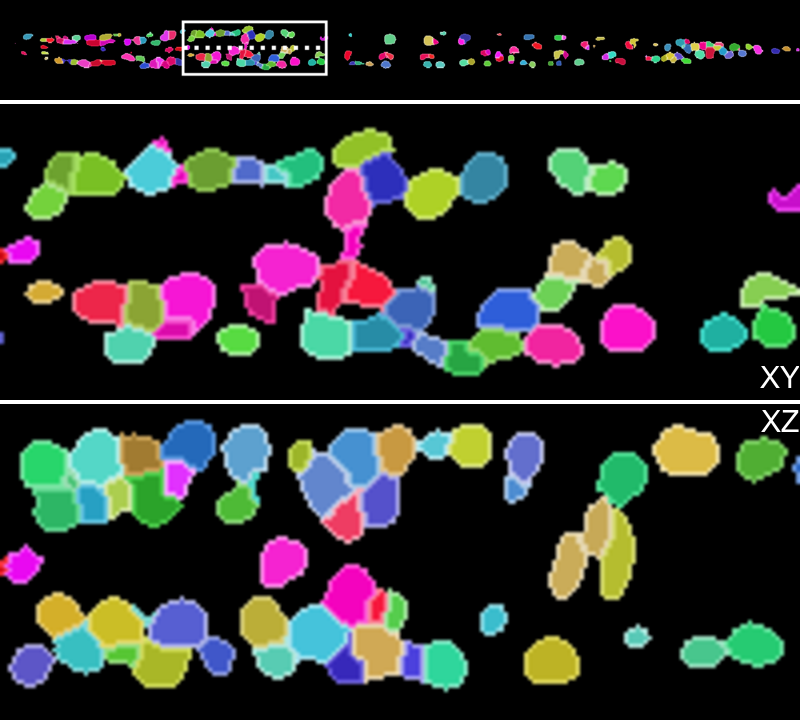}\\
    \textsc{U-Net} & \textsc{U-Net+} & \textsc{Stardist-3D}\\
      \includegraphics[width=.32\linewidth]{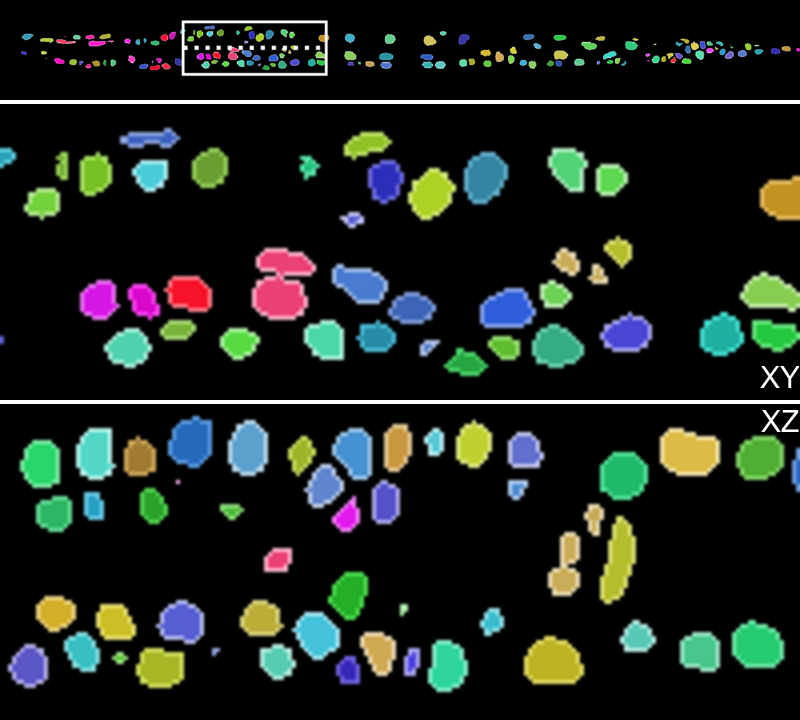}
      &\includegraphics[width=.32\linewidth]{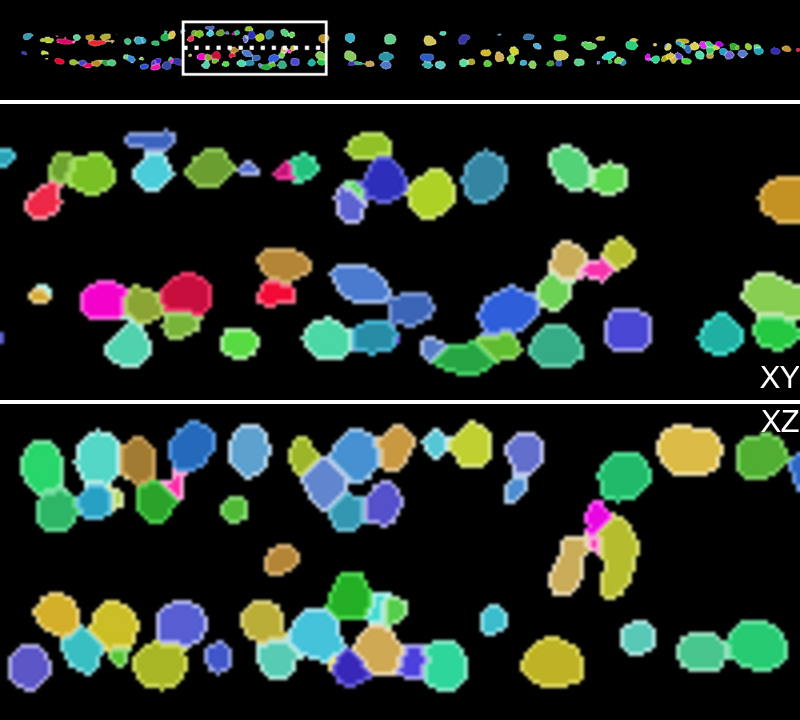}
      &\includegraphics[width=.32\linewidth]{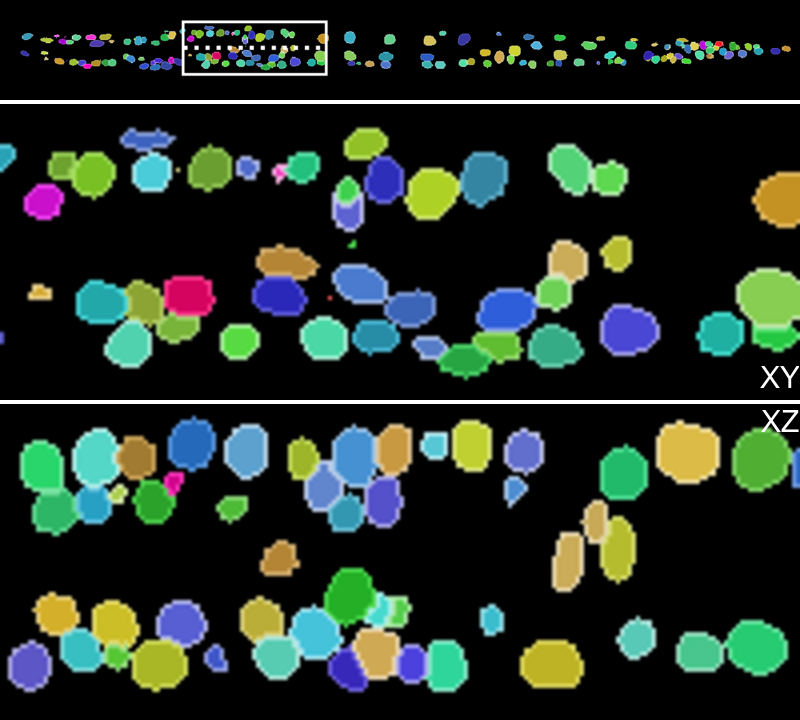}
  \end{tabular}
  \\[1em]
  b) \datako \\
  \begin{tabular}{ccc}
    Input & GT & \textsc{IFT-Watershed}\\
      \includegraphics[width=.32\linewidth]{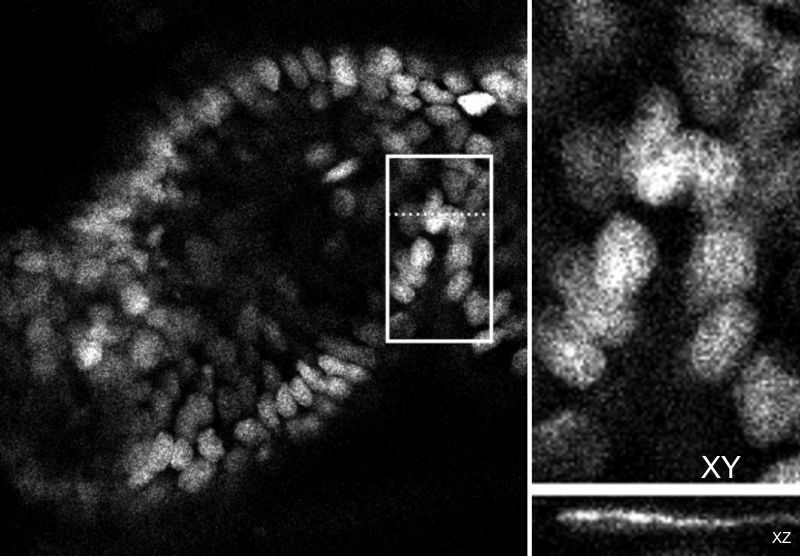}
      &\includegraphics[width=.32\linewidth]{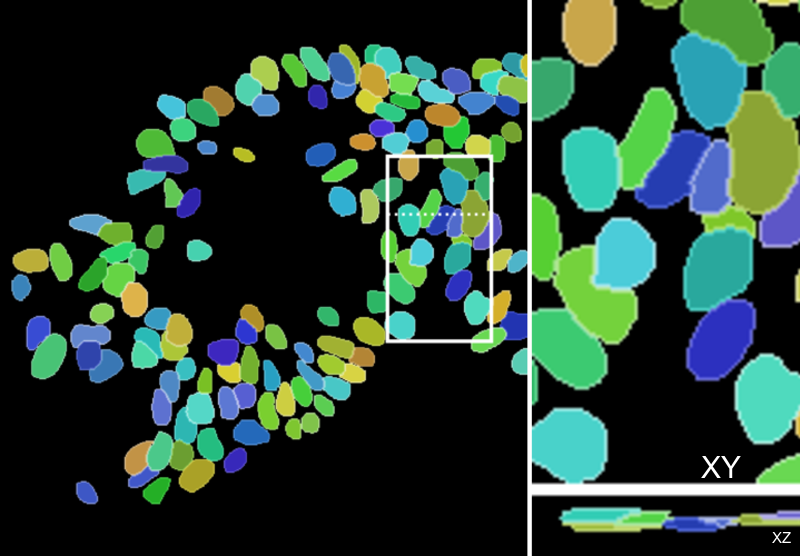}
      &\includegraphics[width=.32\linewidth]{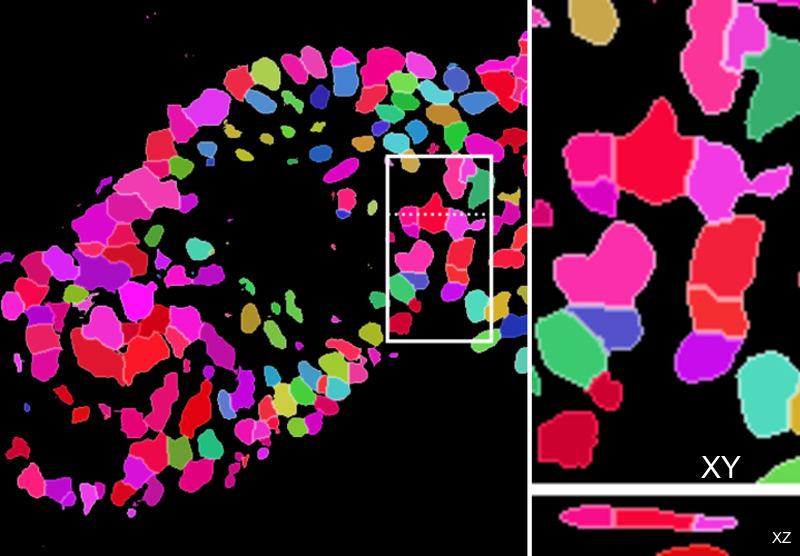}\\
    \textsc{U-Net} & \textsc{U-Net+} & \textsc{Stardist-3D}\\
      \includegraphics[width=.32\linewidth]{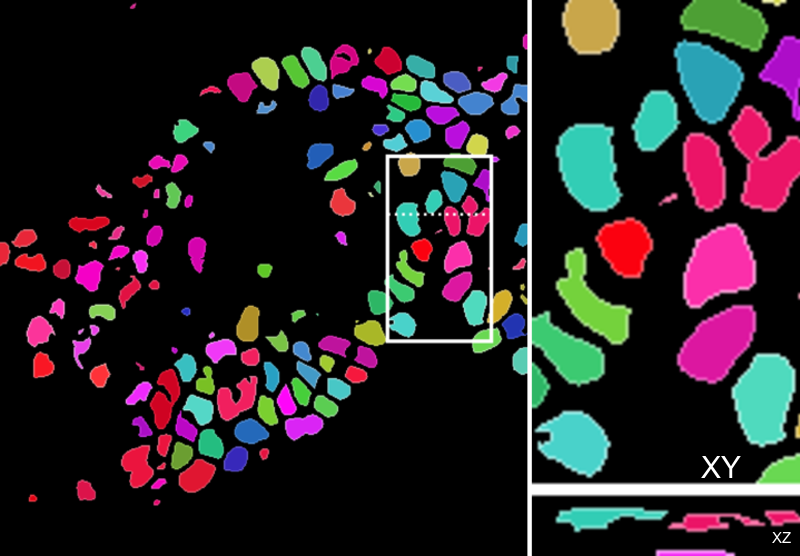}
      &\includegraphics[width=.32\linewidth]{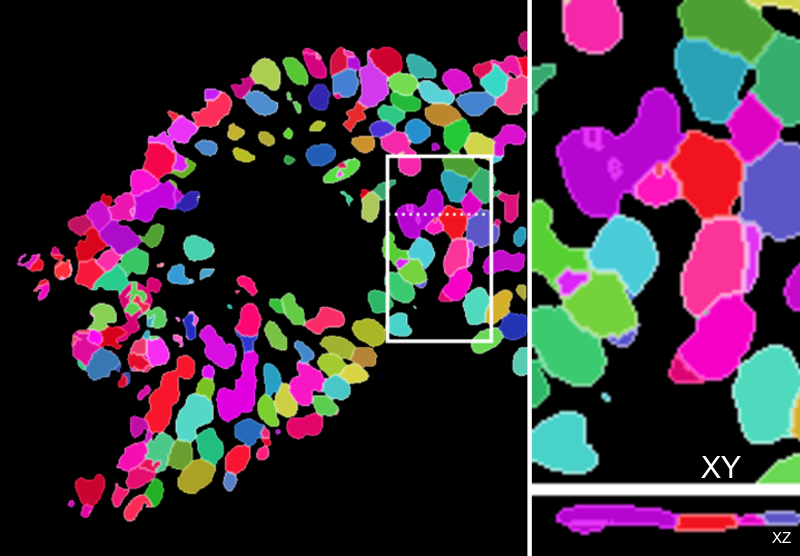}
      &\includegraphics[width=.32\linewidth]{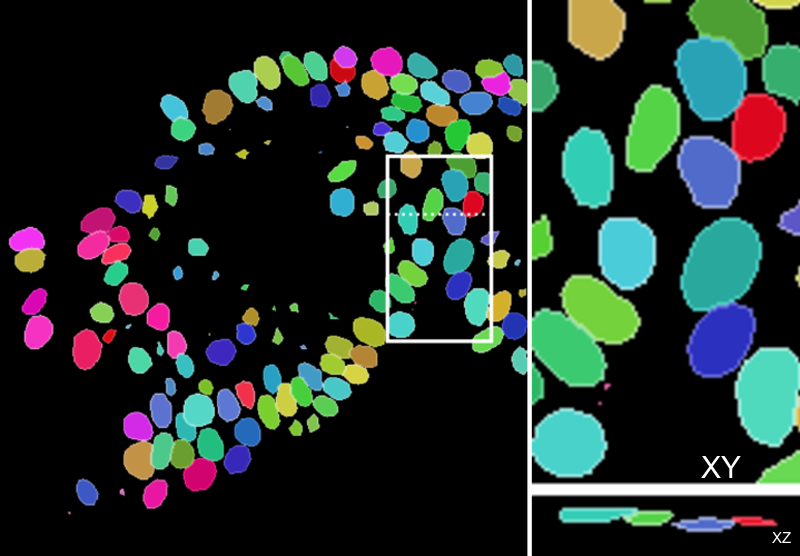}
  \end{tabular}
\caption{%
Example results ($\tau=0.5$) of all methods for both datasets. %
Colors denote nucleus identities, \ie correct predictions ($\mathit{TP}$) have the same color as the ground-truth (GT).
Incorrect predictions ($\mathit{FP}$) are shown in red hues. False negatives ($\mathit{FN}$) are not highlighted. For each inset we show lateral (XY) and axial (XZ, indicated by dotted line) views.%
}
\vspace{12pt}
\label{fig:results}
\end{figure*}
}}

\newcommand{\figAPplot}{{
\begin{figure*}[t!]
  \centering
  \begin{subfigure}[t]{.49\linewidth}
    \begin{overpic}[width=1\linewidth]{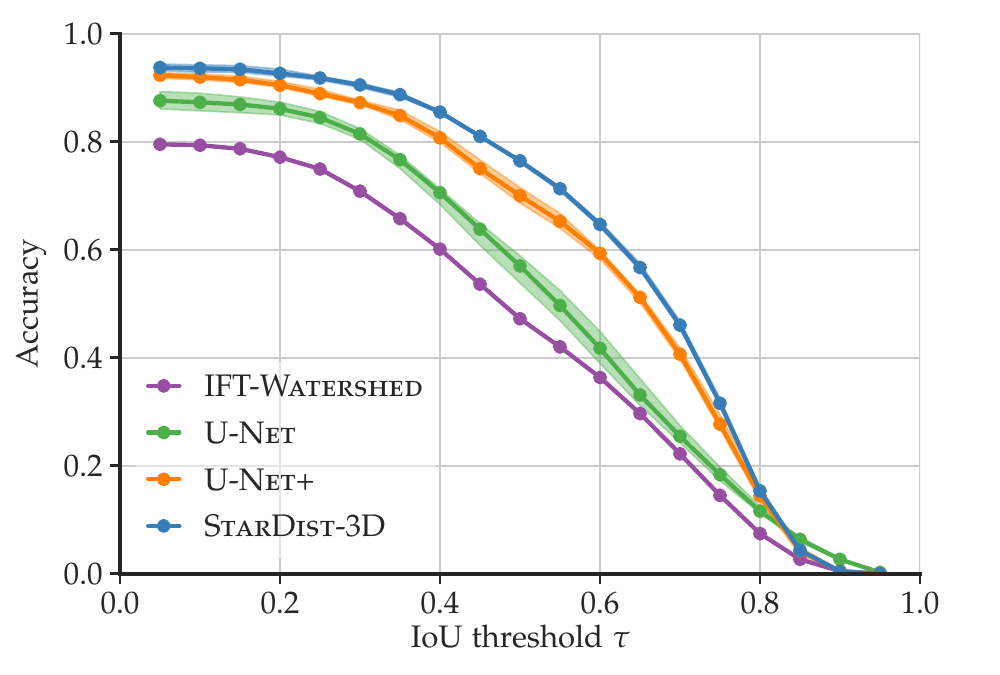}
      \put(-2,62){a)}
    \end{overpic}
  \end{subfigure}
  \begin{subfigure}[t]{.49\linewidth}
    \begin{overpic}[width=1\linewidth]{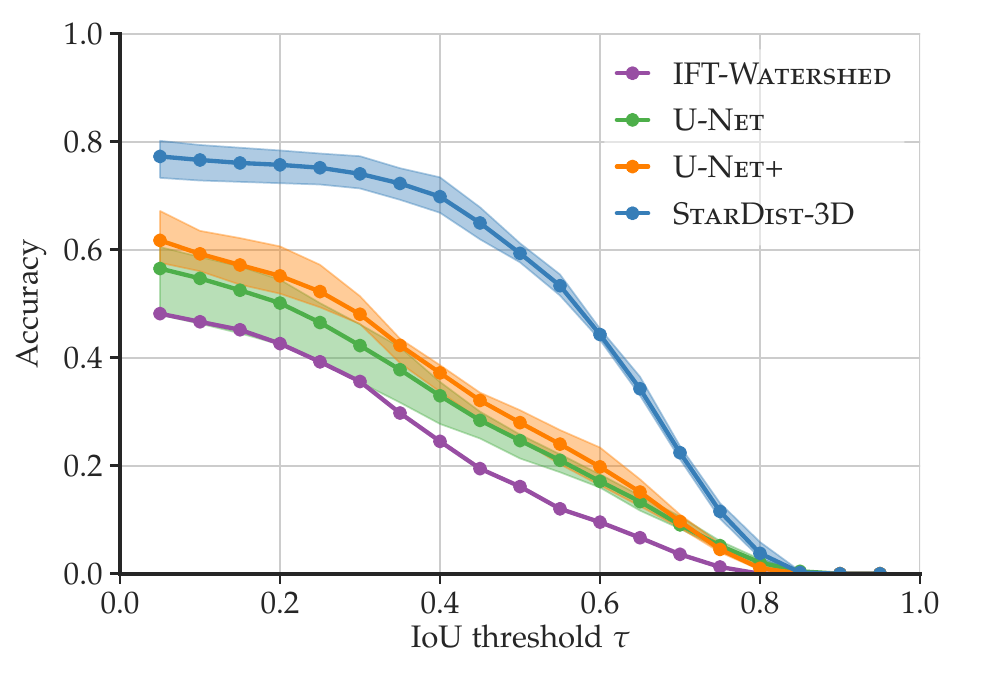}
      \put(-2,62){b)}
    \end{overpic}
  \end{subfigure}
  \vspace{-6pt}
\caption{%
  Accuracy for several IoU thresholds $\tau$ for datasets a) \dataworm and b) \datako.
  We show the average performance over $5$ independent trials for all trained models (shaded regions indicate best and worst result).
}
\label{fig:applot}
\vspace{-3pt}
\end{figure*}
}}

\newcommand{\figRays}{{
\begin{figure}[t]
  \centering
  \includegraphics[width=1\linewidth]{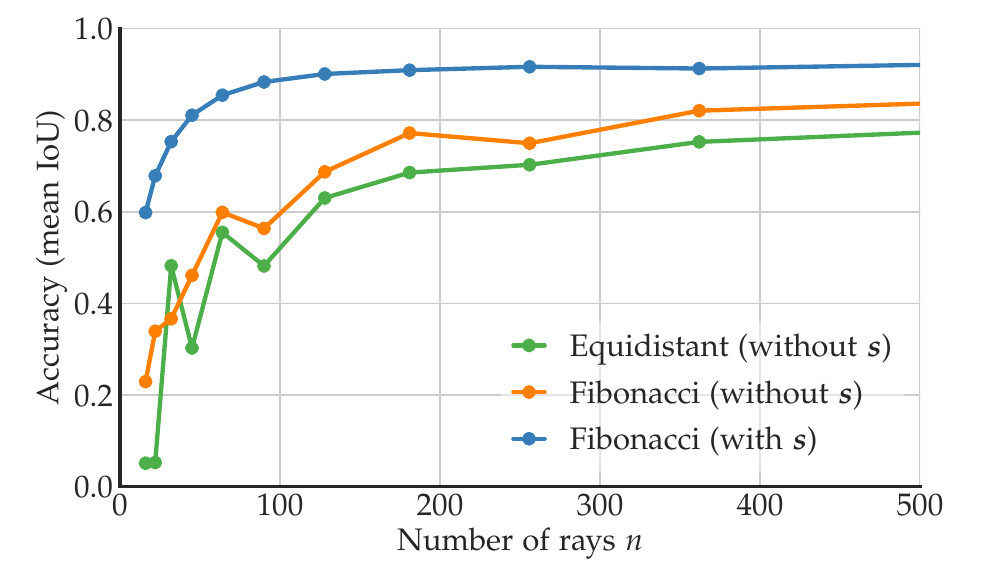}%
  \vspace{-6pt}
  \caption{%
    Reconstruction accuracy (mean intersection over union) of ground-truth instances when using different unit rays (Equidistant/Fibonacci) and anisotropy factors $s$ (for dataset \datako).
  }
\label{fig:rays}
\end{figure}
}}

\newcommand{\figSparse}{{
\begin{figure}[h]
  \centering
  \includegraphics[width=1\linewidth]{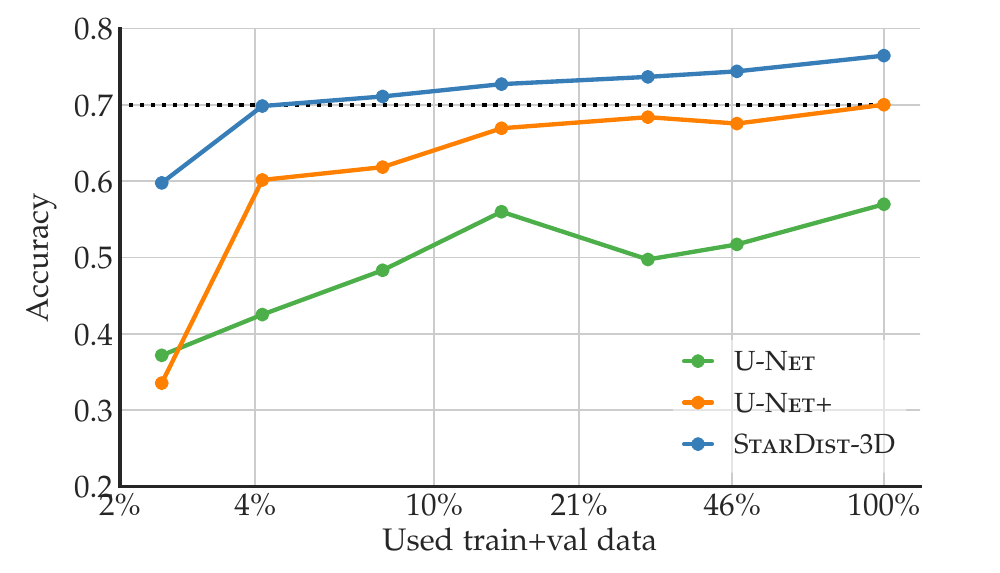}
  \vspace{-18pt}
  \caption{%
    Test accuracy ($\tau=0.5$) of different methods when using only a fraction of all available training/validation volumes (for dataset \dataworm).
  }
\label{fig:sparse}
\end{figure}
}}

\newcommand{\figPaintera}{{
\begin{figure*}[t]
\centering%
\includegraphics[width=.495\linewidth]{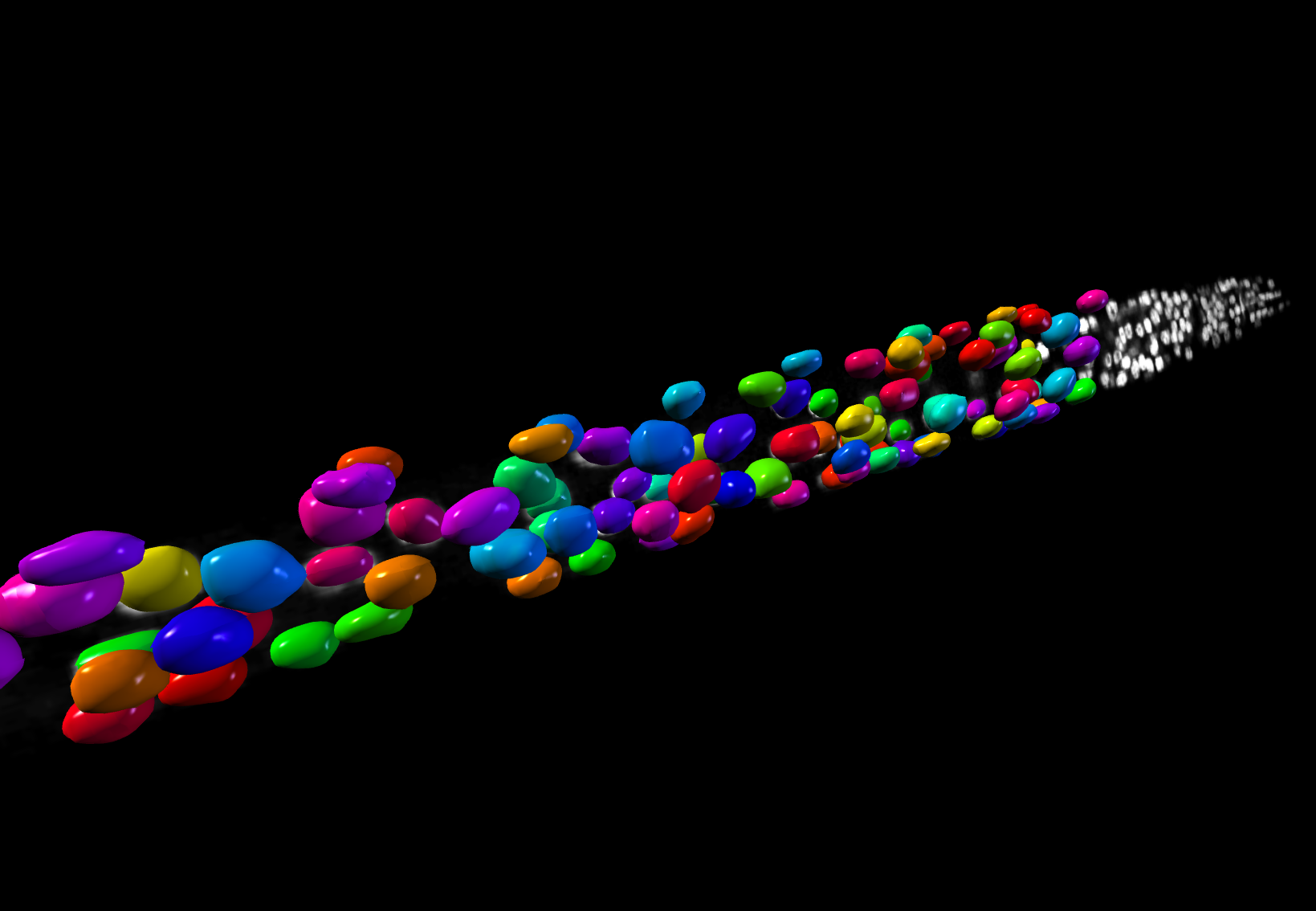}\hfill%
\includegraphics[width=.495\linewidth]{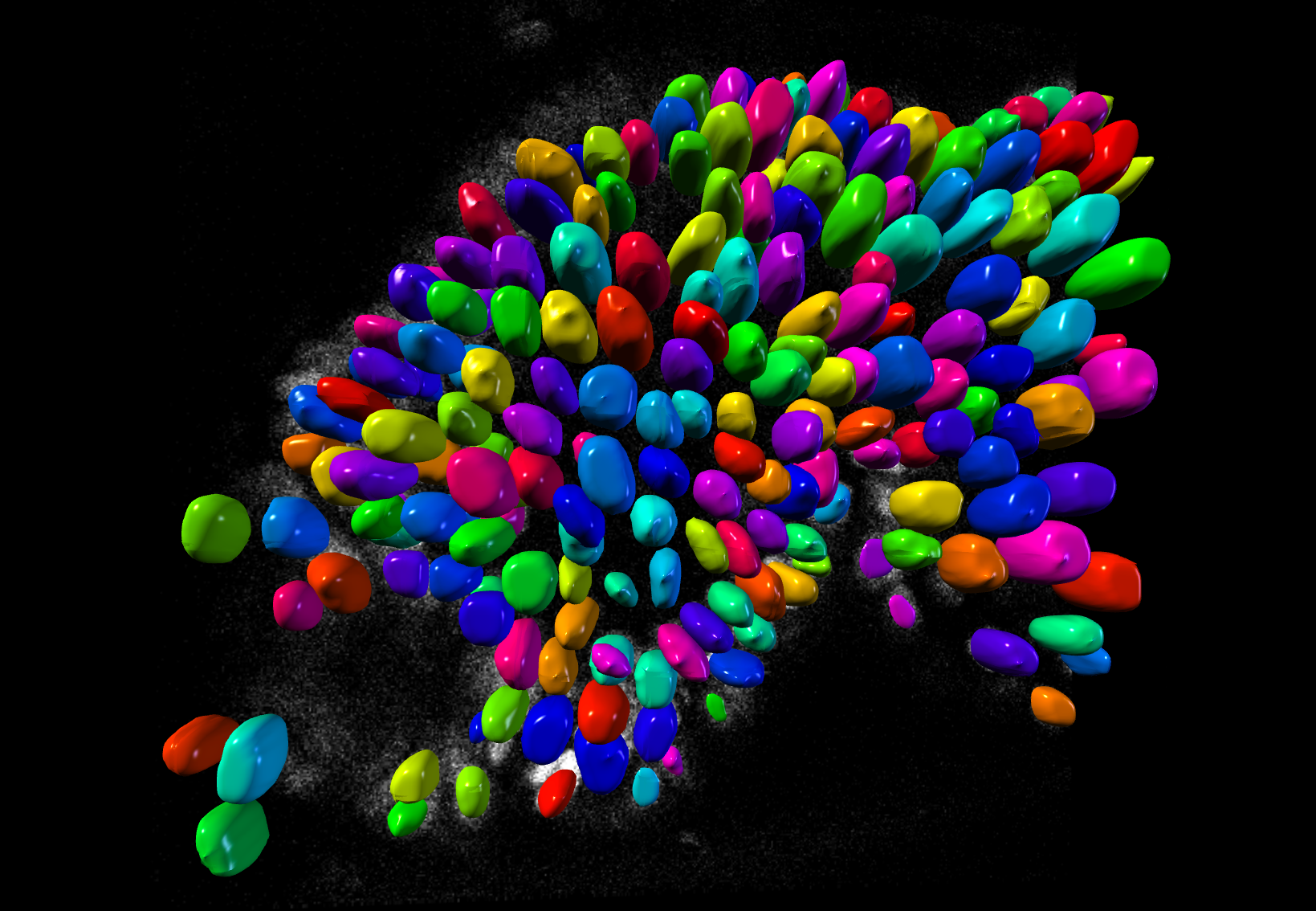}%
\caption{%
Example results of our \stardist approach for both datasets \dataworm (left) and \datako (right). Each instance of a predicted cell nucleus is assigned a random color (not all shown for \dataworm). 3D rendering via \emph{Paintera}~(\url{https://github.com/saalfeldlab/paintera}).
}
\label{fig:paintera}
\end{figure*}
}}

\section{Introduction}

Detection and segmentation of cell nuclei in volumetric (3D) fluorescence microscopy images is a ubiquitous problem in developmental biology and often constitutes the first step when studying cellular expression patterns, or when tracing cell lineages in developing organisms~\cite{meijering2012,ulman2017}.
The task of nuclei \emph{detection} is to roughly locate all individual nuclei inside a 3D volume, \eg by enumerating their center points or bounding boxes.
On the other hand, \emph{semantic segmentation} aims to label each pixel with a semantic class (\eg, nucleus or background), but is not concerned with discerning individual nuclei.
Finally, \emph{instance segmentation} is more ambitious since it combines these tasks by seeking a separate label mask for each individual nucleus.
As modern microscopes produce increasingly large 3D datasets, many automated instance segmentation methods have been proposed over the years~\cite{meijering2012}.
These include classical thresholding approaches with pixel-grouping via connected component, morphological methods based on the watershed transform~\cite{beucher1992,zampirolli2002,cheng2009}, and optimization via graph-cuts~\cite{boykov2006}.
More recently, methods based on deep learning have been shown to vastly improve results for natural and biomedical images alike~\cite{he2017,vanvalen2016,Xie2018}.

\figOverview

In general, deep learning based instance segmentation can be roughly categorized into
\emph{1)} methods that first perform semantic segmentation followed by grouping of pixels into distinct objects (\eg U-Net~\cite{cciccek2016,caicedo2019}), and
\emph{2)} methods that first predict axis-aligned bounding boxes of individual objects with a subsequent semantic segmentation step for each found object (\eg~\cite{he2017,zhao2018,xu2018}).
Despite the advances made by these methods, they often still underperform due to the low signal-to-noise ratios and dense packing of nuclei in typical fluorescence microscopy datasets. %
In particular, methods of category \emph{1)} are prone to erroneously fuse touching nuclei, and those of
category \emph{2)} may fail to discern objects that are poorly approximated with bounding boxes.

These problems have recently been highlighted by Schmidt \etal~\cite{schmidtweigert2018} for the case of 2D fluorescence microscopy images.
To alleviate these issues,~\cite{schmidtweigert2018} proposed a method called {\small\textsc{StarDist}}, which uses a neural network that directly predicts an appropriate shape representation (star-convex polygons) for cell nuclei and demonstrated improved results.
Concretely, for every pixel inside an object (nucleus) they predict the distance to the object boundary along several radial directions, thus defining a star-convex polygon. Furthermore, they also predict an object probability to determine which pixels are part of cell nuclei and thus are allowed to vote for an object shape. Since every pixel is predicting a polygon to represent the entire shape of the object it belongs to, they perform non-maximum suppression to prune redundant shapes that likely represent the same object.
Note that \cite{schmidtweigert2018} sits somewhere in between object detection and instance segmentation
because the predicted shapes are of relatively high fidelity, but are not pixel-accurate.

In this paper, we adopt and extend the approach of~\cite{schmidtweigert2018}
to the case of 3D volumes and use \emph{star-convex polyhedra} as shape
representations for cell nuclei and similar shapes.
We directly
predict the polyhedra parameters densely for each pixel and then use
non-maximum suppression (NMS) to prune the highly redundant set of obtained
polyhedron shapes to ideally retain only one predicted shape for each true
object in the image. Please see \cref{fig:overview} for an overview of our approach.
Note that we keep the benefits of \cite{schmidtweigert2018},
first and foremost to accurately disambiguate densely packed objects in images
with low signal-to-noise ratios. Furthermore,
star-convex polygons/polyhedra are a superset of convex shapes in 2D/3D
and thus include common shapes like bounding boxes and circles/spheres as special cases.

\vspace{-6pt}
\paragraph{Contributions}
The extension of \cite{schmidtweigert2018} from 2D to 3D is challenging
and our main contribution in this paper.
First, computing the intersection of two star-convex polyhedra
(as required for NMS) efficiently is non-trivial (see \cref{sub:nms} and
\cref{fig:overview}c), but highly necessary to make this approach practical
for large 3D volumes.
Second, while \cite{schmidtweigert2018} used $32$ radial directions to
represent 2D nuclei shapes, a naive extension to 3D would require $32^2=1024$
directions which is not feasible due to the excessive amount of
computation and memory required for large 3D volumes.
We show that a more judicious selection of radial directions (\cref{sec:method} and
\cref{fig:overview}a) enables faithful shape representations with as
little as $64$ values. %
Third, microscopy volumes are commonly acquired with anisotropic voxel sizes that result in squeezed nuclei shapes along the axial (Z) direction.
We find that it is critical to
adapt the star-convex representation to account for this anisotropy of the data
to achieve good results (\cref{sec:method,sec:experiments}).
Finally, we demonstrate on two challenging datasets that our proposed method (\stardist) leads to superior results when compared to a classical watershed method and U-Net baselines. %

\section{Method}
\label{sec:method}

\subsection{Star-convex polyhedra}
\label{sec:starconvex}
We describe the 3D shape of a single object (cell nucleus) with a \emph{star-convex polyhedron}.
Concretely, for each pixel inside an object we compute the distances $d_k$ to the object boundary along a fixed set of $n$ \emph{unit rays} $\vec{r}_k$.
To obtain a faithful shape model, we use rays that are approximately evenly distributed on an ellipsoid  representative of the objects in a dataset.
To that end, we first compute the points $(x_k,y_k,z_k)_{k=0\ldots n-1}$ of a spherical \emph{Fibonacci lattice}~\cite{gonzalez2010}
\begin{align*}
  z_k &= -1+\tfrac{2k}{n-1}, \\
  y_k &= \sqrt{1-z_k^2} \sin{\bigl[2\pi (1-\varphi^{-1}) k \bigr]}, \\
  x_k &= \sqrt{1-z_k^2} \cos{\bigl[ 2\pi (1-\varphi^{-1}) k \bigr]},
\end{align*}
where $\varphi = \tfrac{1+\sqrt{5}}{2}$ denotes the golden ratio.
To account for anisotropy of the data we generate intermediate, anisotropically scaled vectors $\vec{u}_k = \bigl(\tfrac{x_k}{s_x},\tfrac{y_k}{s_y},\tfrac{z_k}{s_z}\bigr)$.
The respective \emph{anisotropy factor} $\vec{s} = (s_x, s_y, s_z)$
is calculated as the median bounding box size of all objects in the training images.
The final unit rays $\vec{r}_k$ are then computed via normalization $\vec{r}_k = \frac{\vec{u_k}}{|\vec{u_k}|}$.
The surface of a star-convex polyhedron represented by the distances $d_k$ is then given by its vertices $d_k \cdot \vec{r}_k$ and triangulation induced by the convex hull facets of the unit rays $\vec{r}_k$ (which is a convex set by definition).
\figRays
We generally find that a sufficiently accurate reconstruction of the labeled 3D cell nuclei in our ground-truth (GT) images can be obtained with as few as $64$ rays.
\cref{fig:rays} shows the reconstruction fidelity for a dataset with highly anisotropic images (\datako, \cf~\cref{sec:experiments})
and highlights the importance of using an appropriate anisotropy factor $\vec{s}$.
Note that $\vec{s}$ is automatically computed from the GT images and
does not have to be chosen manually.
Furthermore, \cref{fig:rays} shows that our ray definition (Fibonacci) is more
accurate than using equidistant (polar/azimuthal) distributed rays.

\subsection{Model}
\label{sub:model}

Following \cite{schmidtweigert2018},
we use a convolutional neural network (CNN) to densely predict the star-convex polyhedron representation
and a value that indicates how likely a pixel is part of an object.
Concretely, for each pixel $(x,y,z)$, the CNN is trained to predict the $n$ \emph{radial distances} $\{d_k(x,y,z)\}_{k= 0\ldots n-1}$ to the object boundary as defined above and additionally an \emph{object probability} $p(x,y,z)$ defined as the (normalized) Euclidean distance to the nearest background pixel~(\cref{fig:overview}a).
To save computation and memory we predict at a \emph{grid} of lower spatial resolution than the input image, since a dense (\ie, per input pixel) output is often not necessary (this is similar to the concept of bounding box \emph{anchors} in object detection approaches~\cite{ren2015, redmon2016}).

We use a slightly modified 3D variant of ResNet \cite{he2016} %
as a neural network backbone%
\footnote{We find that using a U-Net~\cite{cciccek2016} backbone leads to very similar results.}
to predict both the radial distances and object probabilities
(\cref{fig:overview}b).
In particular, we use residual blocks with $3$ convolution layers of kernel size \sizethree{3}{3}{3}. %
Similar to \cite{he2016}, we start with two convolution layers of kernel sizes \sizethree{7}{7}{7} and \sizethree{3}{3}{3}, but without strides to avoid downsampling. This is followed by $m$ residual blocks, where
each block
only performs downsampling %
if the spatial resolution is still higher than the %
prediction grid (see above); we double the number of convolution filters after each downsampling.
After the last residual block, we use a single-channel convolution layer with \emph{sigmoid} activation to output the per-pixel\footnote{To improve readability we will drop from now on the explicit pixel coordinate $(x,y,z)$ for both $p(x,y,z)$ and $d_k(x,y,z)$.} object probabilities $p$.
The last residual block is additionally %
connected to an $n$-channel convolution layer to output the radial distances $d_k$.
Our code based on Keras/TensorFlow~\cite{chollet2015,abadi2016} and documentation is available at \url{https://github.com/mpicbg-csbd/stardist}.

\paragraph{Training}
Given the pixel-wise object probabilities and distances of the prediction $(\hat{p}, \hat{d}_k)$ and ground-truth  $(p, d_k)$, we minimize the following loss function (averaged over all pixels) during training:
\begin{equation}
  L(p, \hat{p}, d_k, \hat{d}_k) =  L_{\mathit{obj}}(p, \hat{p}) + \lambda_d L_{\mathit{dist}}(p, \hat{p}, d_k, \hat{d}_k).
  \label{eq:loss}
\end{equation}
For the object loss $L_{\mathit{obj}}$ we use standard \emph{binary cross-entropy}
\begin{equation}
  L_{\mathit{obj}}(p, \hat{p}) = -p \log \hat{p} - (1-p) \log (1-\hat{p}).
  \label{eq:loss_obj}
\end{equation}
For the distance loss $L_{\mathit{dist}}$ we use the \emph{mean absolute error} weighted by the object probability (active only on object pixels, \ie $p>0$) and add a regularization term (active only on background pixels, \ie $p=0$):
\begin{multline}
  L_{\mathit{dist}}(p, \hat{p}, d_k, \hat{d}_k) =  p \cdot \mathbbm{1}_{p>0} \cdot \frac{1}{n}\sum\nolimits_k    |  d_k -\hat{d}_k| +\\
\lambda_{\mathit{reg}} \cdot \mathbbm{1}_{p=0} \cdot  \frac{1}{n}\sum\nolimits_k   |\hat{d}_k|.
  \label{eq:loss_dist}
\end{multline}
This specific form was chosen to promote increased accuracy for points closer to the object centers (which eventually will be used as polyhedra center candidates).

\figDatasets

\paragraph{Prediction}
After the CNN predicts the
radial distances $\hat{d}_k$ and object probabilities $\hat{p}$,
we collect a set of object \emph{candidates}
by only considering radial distances at pixels with object probabilities above a reasonably high threshold, \ie we only retain shapes that very likely belong to an object. %
Since the set of object candidates is highly redundant,
we use non-maximum suppression (NMS) to obtain only one shape for every actual object in the image, as is common in object detection (\eg, \cite{he2017}).
Thereby, the object candidate
with the highest object probability suppresses all other remaining
candidates if they overlap substantially. %
This process is repeated until there are no further candidates to be suppressed.
All remaining (\ie not suppressed) candidates yield the final set of predicted object shapes.

\subsection{Efficient non-maximum suppression}
\label{sub:nms}

The NMS step requires to assess the pairwise overlap of a potentially large set of polyhedron candidates ($> 10^4$).
Unfortunately, computing the exact intersection volume between two star-convex polyhedra \emph{efficiently} is non-trivial (in contrast to \emph{convex} polyhedra).
To address this issue, we employ a filtering scheme that computes as needed successively tighter upper and lower bounds for the overlap of polyhedron pairs~(\cf~\cref{fig:overview}c).
Concretely, we compute the intersection volume of the respective \emph{i)} bounding spheres (upper bound), \emph{ii)} inscribed spheres (lower bound), \emph{iii)} convex hulls (upper bound), and \emph{iv)} \emph{kernels}\footnote{The (convex) set of all points that can serve as center of the star-convex polyhedron.} (lower bound).
Note that \emph{iii)} and \emph{iv)} involve only the intersection of convex polyhedra and can thus be computed efficiently~\cite{barber1996}.
If a computed bound is already sufficient to decide whether a candidate should be suppressed or not,
no further computation is carried out.
Otherwise, we eventually perform the expensive but exact intersection computation by rasterization of both polyhedra.
We find that this NMS filtering scheme leads to a noticeable reduction in runtime  that makes \stardist practical (\eg $9$~s for a stack of size $\sizethree{1141}{140}{140}$ with $12000$ initial candidates).

\figAPplot
\section{Experiments}
\label{sec:experiments}

We consider two qualitatively different
datasets (\cref{fig:datasets})
to validate the efficacy of our approach:
\begin{description}
\item[\normalfont\dataworm]
A subset of $28$ images used in Long \etal~\cite{long2009}, showing
DAPI-stained nuclei of the first larval stage (L1) of \emph{C.~elegans}~(\cref{fig:datasets} left).
Stacks are of average size $\sizethree{1157}{140}{140}$ pixels
with semi-automatically annotated cell nucleus instances ($15148$ in total)
that underwent subsequent manual curation.
We randomly choose $18$/$3$/$7$ images for training/validation/testing.
Note that the images have (near) isotropic resolution.
\item[\normalfont\datako]
A subset of recording \#04 of Alwes \etal~\cite{alwes2016},
showing %
\emph{Parhyale hawaiensis} expressing Histone-EGFP~(\cref{fig:datasets} right).
It contains $6$ images of $\sizethree{512}{512}{34}$ pixels
with manually annotated nucleus instances ($1738$ in total).
We randomly choose $3$/$1$/$2$ images for training/validation/testing.
In contrast to \dataworm, the images are highly anisotropic
in the axial direction. %
This dataset is more challenging due its substantially lower signal-to-noise ratio (\cf~\cref{fig:results}).
Furthermore, it contains much fewer labeled training images,
more akin to what is typical in many biological datasets.
\end{description}

\subsection{Methods and Evaluation}
We compare our proposed \stardist approach against two kinds of methods
(IFT-Watershed~\cite{zampirolli2002} and 3D U-Net~\cite{cciccek2016})
that are commonly used for segmentation of fluorescence
microscopy images.
First, a classical watershed-based method that does not use machine learning.
Second, a variant of the popular U-Net with and without more sophisticated postprocessing.

To evaluate the performance of all methods, we use $\mathit{accuracy(\tau)} = \frac{\mathit{TP}}{\mathit{TP}+\mathit{FN}+\mathit{FP}}$ for several overlap thresholds $\tau$.
$\mathit{TP}$ are true positives, which are pairs of predicted and ground-truth nuclei having
an \emph{intersection over union} (IoU) value $\geq\tau$.
$\mathit{FP}$ are false positives (unmatched predicted instances) and
$\mathit{FN}$ are false negatives (unmatched ground-truth instances).
We use the Hungarian method~\cite{kuhn1955hungarian} (implementation from \emph{SciPy}~\cite{scipy}) to compute an optimal matching whereby a single predicted nucleus
cannot be assigned to multiple ground-truth instances (and vice versa).
Note that a suitable value of $\tau$ depends on the biological application. For example, one would likely use a smaller $\tau < 0.5$ for the purpose of counting objects, whereas intensity measurements inside each object would require more accurate shapes and thus demand a higher value of $\tau$.

\begin{description}
\item[\normalfont \stardist]
We use \stardist as explained in \cref{sub:model}
with $n=96$ radial directions and
$m = 3$ residual blocks that
start with $32$ convolution filters.
We predict at a grid half the spatial resolution of the input image,
except for the anisotropic Z axis of \datako. %
We use automatically computed anisotropy factors (\cf~\cref{sec:starconvex}) of $\vec{s} = (1,1,1)$ for \dataworm
and $\vec{s} = (1,1,7.1)$ for \datako.
We use weights $\lambda_d = 0.1$ and $\lambda_{reg} = 10^{-4}$ for the loss function in ~\cref{eq:loss}.

\item[\normalfont \hmaxima]

The IFT-Watershed~\cite{zampirolli2002} is an %
efficient combination of maxima detection and 3D watershed segmentation.
It represents an advanced classical image segmentation method
that we know is being used in practice.
Concretely, we use the \emph{Interactive Watershed} plugin\footnote{\url{https://imagej.net/Interactive_Watershed}} %
in \emph{Fiji}~\cite{Schindelin2012}
and perform extensive parameter tuning (such as Gaussian filter size during preprocessing and maxima detection thresholds) using the training images of each dataset.

\item[\normalfont \unet]
We train a 3D U-Net \cite{cciccek2016} to classify each pixel into \emph{background}, \emph{nucleus}, and also \emph{nucleus boundary}, as this helps substantially to separate touching nuclei~\cite{caicedo2019}. %
We threshold the predicted \emph{nucleus} probabilities and group pixels in each connected component to obtain individual nuclei. %

\item[\normalfont \unetws]
We use the same trained 3D U-Net as above, but apply more sophisticated postprocessing.
Concretely, we observe improved performance by thresholding the \emph{nucleus} probabilities to obtain seed regions that we grow (via 3D watershed~\cite{scikit-image}) until they reach pixels with \emph{background} probability above a second threshold. %

\end{description}
We apply random data augmentations during training, including %
flips, axis-aligned rotations, elastic deformations, %
intensity rescaling, and noise.
After training, thresholds for all methods (as described above)
are tuned on validation images to optimize accuracy averaged over $\tau \in \{0.3,0.5,0.7\}$.

\begin{table*}[t]
  \centering
  \begin{tabular}{@{} l S[table-format=1.4]S[table-format=1.3]S[table-format=1.3]S[table-format=1.3]S[table-format=1.3]S[table-format=1.3]S[table-format=1.3]S[table-format=1.3]S[table-format=1.3] @{}}
    \toprule
    Threshold $\tau$ & {0.10} & {0.20} & {0.30} & {0.40} & {0.50} & {0.60} & {0.70} & {0.80} & {0.90}\\
    \midrule
    \multicolumn{10}{c}{{\dataworm}}\\
    \midrule
    \textsc{IFT-Watershed}  & 0.794 & 0.771 & 0.708 & 0.601 & 0.472 & 0.364 & 0.222 & 0.074 & 0.005 \\
    \textsc{U-Net}                  & 0.873 & 0.861 & 0.814 & 0.706 & 0.570 & 0.418 & 0.255 & 0.116 & \bfseries 0.027 \\
    \textsc{U-Net+}               & 0.920 & 0.905 & 0.872 & 0.807 & 0.700 & 0.593 & 0.406 & 0.144 & 0.005 \\
    \stardist            & \bfseries 0.936 & \bfseries 0.926 & \bfseries 0.905 & \bfseries 0.855 & \bfseries 0.765 & \bfseries 0.647 & \bfseries 0.460 & \bfseries 0.154 & 0.004 \\
    \midrule
    \multicolumn{10}{c}{{\datako}}\\
    \midrule
    \textsc{IFT-Watershed}  & 0.467 & 0.426 & 0.356 & 0.245 & 0.161 & 0.096 & 0.036 & 0.000 &  0.000 \\
    \textsc{U-Net}                  & 0.547 & 0.501 & 0.423 & 0.330 & 0.247 & 0.171 & 0.091 & 0.021 & 0.000 \\
    \textsc{U-Net+}               & 0.592 & 0.552 & 0.481 & 0.372 & 0.280 & 0.198 & 0.097 & 0.010 & 0.000 \\
    \stardist             & \bfseries 0.766 & \bfseries 0.757 & \bfseries 0.741 & \bfseries 0.698 & \bfseries 0.593 & \bfseries 0.443 & \bfseries 0.224 & \bfseries 0.038 & 0.000 \\
    \bottomrule
  \end{tabular}
  \vspace{-6pt}
  \caption{Accuracy (average over $5$ independent trials for trained models) for several IoU thresholds $\tau$ for datasets \dataworm and \datako.
}
  \label{tab:results}
  \vspace{-2pt}
\end{table*}

\subsection{Results}

The results in \cref{tab:results,fig:applot} show that \stardist
consistently outperforms all other methods that we compared to
(note that we report the average result over $5$ trials for all trained models).
The performance gap between \stardist and the other methods is especially
striking for dataset \datako, which may be explained by \stardist's shape
model being especially helpful to disambiguate between neighboring nuclei
in these challenging low-SNR images.
In~\cref{fig:results} we show lateral (XY) and axial (XZ) views of segmentation results for both datasets.
Here, \hmaxima often produces imperfect boundaries and erroneous splits, particularly for dataset \datako. This is expected, as the watershed operation uses the input intensities alone without leveraging extracted features.
\unet tends to under-segment the image, generally producing object instances that are too small, as the use of a single threshold for the \emph{nucleus} class leads to a trade-off between object size and avoidance of falsely merged objects.
In contrast, \unetws exhibits slight over-segmentation, since a larger first threshold produces more (but smaller) objects that are then grown to yield the final instances.
Finally, \stardist produces superior segmentations, although it can sometimes fail to detect some nuclei (especially for dataset \datako). As an additional visualization, we show a 3D rendering of \stardist segmentation results for both datasets in~\cref{fig:paintera}.

Note that we find (not shown) that the accuracy of \stardist
would drop dramatically (for example, from $0.593$ to $0.291$ for $\tau=0.5$)
if we did not adapt the radial directions to account for the anisotropy of the
nuclei shapes (\cref{sec:starconvex}) for \datako.
While \stardist's lead is less pronounced for dataset \dataworm, this may be
due to the higher signal quality of the input images and also the general
abundance of labeled cell nuclei available for training and validation
($11387$ in total). %
In \cref{fig:sparse}, we investigate how \stardist and the other trained
models cope with less annotated data by randomly selecting only
a partial 3D image slice from each training and validation stack.
Interestingly, we find that with only $4.15\%$ of the training and validation data
($472$ instances in total), \stardist can for $\tau=0.5$ reach the same performance
(accuracy of $0.7$)
as \unetws with access to $100\%$ of the
training and validation data.

\section{Discussion}

We presented \stardist, an extension of \cite{schmidtweigert2018} to detect
and segment cell nuclei in volumetric fluorescence microscopy images,
even when they exhibit substantial anisotropy.
Our method outperformed strong watershed and U-Net baselines, yet is easy to train and use,
and due to our star-convex polyhedra parameterization and efficient
intersection implementation fast enough to process typical large 3D volumes.
Furthermore, \stardist should be generally applicable to segment objects whose
shapes are well-represented with star-convex polyhedra.

\subsubsection*{Acknowledgments}

We thank Frederike Alwes and Michalis Averof (IGFL) for providing Parhyale data, Dagmar Kainmüller (MDC Berlin) for worm annotation and curation, and Stephan Saalfeld and Philipp Hanslovsky (HHMI Janelia) for software assistance.
Uwe Schmidt and Robert Haase were supported by the BMBF grant \emph{SYSBIO II} (031L0044),
and Ko Sugawara by the ERC Advanced Grant \emph{reLIVE} (ERC-2015-AdG-694918).

\figSparse
\figResults
\figPaintera

{\small
\bibliographystyle{ieee}

}

\end{document}